# Retrieving and ranking short medical questions with two stages neural matching model

Xiang Li
School of Computer Science
University of Nottingham,
Ningbo, China
Xiang.Li@nottingham.edu.cn

Xinyu Fu
School of Computer Science
University of Nottingham
Ningbo, China
Xinyu-Fu@nottingham.edu.cn

Zheng Lu*
School of Computer Science
University of Nottingham
Ningbo, China
Zheng.Lu@nottingham.edu.cn

Ruibin Bai
School of Computer Science
University of Nottingham
Ningbo,China
Ruibin.Bai@nottingham.edu.cn

Uwe Aickelin
School of Computing and
Information Systems
University of Melbourne
Melbourne,Australia
uwe.aickelin@unimelb.edu.au

Peiming Ge
Technology Dept
Ping An Health Cloud
Ping An, Shanghai
gepeiming649@jk.cn

Gong Liu
Technology Dept
Ping An Health Cloud
Ping An, Shanghai
liugong189@jk.cn

*Abstract*—Internet hospital is a rising business thanks to recent advances in mobile web technology and high demand of health care services. Online medical services become increasingly popular and active. According to US data in 2018, 80 percent of internet users have asked health-related questions online. Numerous data is generated in unprecedented speed and scale. Those representative questions and answers in medical fields are valuable raw data sources for medical data mining. Automated machine interpretation on those sheer amount of data gives an opportunity to assist doctors to answer frequently asked medical-related questions from the perspective of information retrieval and machine learning approaches. In this work, we propose a novel two-stage framework for the semantic matching of query-level medical questions, which takes advantages of sentence similarity-based search engine techniques and Siamese inspired recent recurrent neural network. The two-stage hierarchical design optimises the performance of automatic information retrieval of user queries. Compared against the classical TFIDF search technique as a single-stage, our novel soft search technique performs significantly better. Incorporating an advanced deep learning model as the second stage can improve the results further, which we believe is the new state-of-the-art in the current problem setting with the unique medical corpus from one of the largest online healthcare provider in market.

## I. INTRODUCTION

Online medical services are new business at the present era of fast developing mobile web technology and high demand of health care services. Major portion of internet users have sought to find health care answers online. Data is produced in a booming scale in recently years. Computerised techniques to fetch and mine on those high volume of medical query streams have attracted much attention in recent years. Machine-aided approaches exert positive influences on daily medical practice and reduce replicated labor work for doctors. That motivates us to develop a robust two-stage neural matching model for automated analysis of user queries against the standard question pool.

The proposed two-stage neural matching framework has been successfully deployed to the one of leading medical mobile platform with on-average daily active user over 0.3 million. The typical scenario of the proposed framework is that when a user or patient raises a question about her/his illness or medicine intake, our model is triggered to first traverse through the frequently asked questions (FAQ) list and gives confidence score on the top-n matching questions. The final confidence score-ranking list is obtained from the second step deep neural network once the first step search engine has generated the intermediate result. High confidence score usually implies high relevance of the patient query matching the standard FAQ list. By doing so, online doctors' routine work can be eased to some extent, which saves money from the company perspective.

Semantic interpretation on human generated texts has long been a difficult task in natural language processing (NLP). The applications of natural language semantic understanding range from machine translation, document retrieval, and question and answering (Q&A). Formally speaking, the definition of matching two given texts $T_1 = (\omega_1, \omega_2, \cdots, \omega_m)$ and $T_2 = (v_1, v_2, \cdots, v_n)$ is represented as a confidence score by the match function below:

$$\text{match}(T_1, T_2) = F(\phi(T_1), \phi(T_2)) \quad (1)$$

where $\omega_i$ and $v_j$ denote the i-th and j-th word in $T_1$ and $T_2$ respectively. For the particular language, $\omega_i$ and $v_j$

denote the i-th and j-th phrase after segmentation as some language text preserves no space in between. The match function applies to any languages since word segmentation techniques are universally applicable. Thus our framework proposed in the current paper is language independent.

Matching score identification requires a strong textual snippets semantic representation schema and sentence similarity calculation approach [1, 2]. More specifically, rich representation of the hidden structures of texts and precise modelling of the interaction of the two sentences are the keys to determine sentence similarity. We simplify our collaborator's FAQ scenario as a duplicate sentence pair comparison problem, given that an expert maintained FAQ list is fixed. To the best of our knowledge, the FAQ list is uniquely provided and expresses the collective intelligence of a team of experienced doctors on general medical practices.

By investigating in the previous work and relevant literatures, we ought to say that few existing models and algorithms coincide with the novel two-stage framework proposed in the present research. Several successful information retrieval-based models like SMART [3] and page ranking algorithms [4] provide promising results on explicit search queries. Many deep neural networks show competitive performance on tailored duplicate question pair streams. Our two-stage framework makes the most out of each and shows promising effectiveness on the real world practice.

The main contributions of the paper are twofolds: first, we apply our proposed two-stage framework to online medical scenarios that provides a holistic solution from queries to answers. Second, we develop a model that fits the large-scale industry practice with high quality which has promising business values and potential.

The structure of this paper is as follows. Section II provides a review of relevant work. Section III describes our two-stage framework in details. Section IV and Section V present our experiment settings and experiment results respectively. Section VI concludes the current work.

## II. RELATED WORK

Sentence pair matching problem has attracted numerous attentions in the past few decades. Generally speaking, existing methods can be classified into non-neural network approaches and neural network-based frameworks. In the recent days and near future, neural network-related methods become even more popular and due to the fast advances of computing hardware and deep learning software packages.

*A. Non-neural network*

In the early years of paired question classification research, most approaches try to solve the problem by semantic matching. Due to the nature of natural language on diversity, ambiguities, synonyms, and antonyms, simple bag-of-word (BOW) sentence representation tends to have poor performance on semantic question pair classification [5, 6]. As a result, methods such as deeper semantic analysis [7], tree edit-distance [8], and quasi-synchronous grammars [9] have been deployed, which attempt to address the problem by better exploitation of syntactic and semantic structures. However, all aforementioned approaches encounter the barriers of high computational resources consumption and difficulties to employ syntactic or semantic parsers for every pair of sentences or queries.

Learning to rank is another field links to our paired question classification problem, which overlaps with our fine-grained ranking in the present work[10]. Under learning to rank circumstance, ranking is performed based on interpretation between user query and standard question. The standard question pool is often constructed to express the expert knowledge base [11] . Question answering system sheds some light on paired question classification problem since the paired information interaction is common in both [12]. For instance, [13, 14] develop QA models based on medical expert knowledge base. Those previously mentioned frameworks have limited application on medical diagnosis in our proposed duplicate question classification problem.

*B. Neural network*

Neural networks-oriented deep learning models in particular have attracted incredible attentions in recent years and have been successfully applied to the field of natural language processing as well as the specific problem of question-answer pairs matching. Convolutional neural network is adapted to sentence pairs by many researchers [15-17]. As counter parts, recurrent neural network has been applied to multiple works [18, 19]. More deep semantic understanding of natural texts are introduced in the past few years due to the hardware boosting [20]. Hu [21] states a novel multi-model deep belief model to learn hidden semantic information in the textual snippets. Their model incorporates handcrafted features with the traditional *n-gram* features to deliver an enhanced effectiveness. Since the question length in the proposed work is often short compared to the plain data streams, the efficacy of Hu [21]' algorithm is restricted without further enhancement. Our proposed two-stage framework leverages soft cosine similarity on semantic information extraction and deep neural network's competency upon paired information interpretation.

Other work for the paired query classification problem includes recurrent neural network (RNN) and its variants like long short time memory (LSTM) [19, 22, 23]. Grid-level similarity matrix approach is widely applied in the recent investigations on the duplicate question pair problem [19, 24-26]. Technically speaking, hidden neurons for the paired questions can be directly utilised to compute the similarity score or likelihood of the participated two pieces of text [19]. Xiong [27] suggests using attention mechanism to further explore and express low-level information preserved in the paired query with exponentially high computational complexity. We implemented a similar hidden state neural

National Natural Science Foundation of China (Grant No. 71471092), Natural Science Foundation of Zhejiang Province (Grant No. LR17G010001) and Ningbo Municipal Bureau of Science and Technology (Grant No. 2014A35006, 2017D10034)

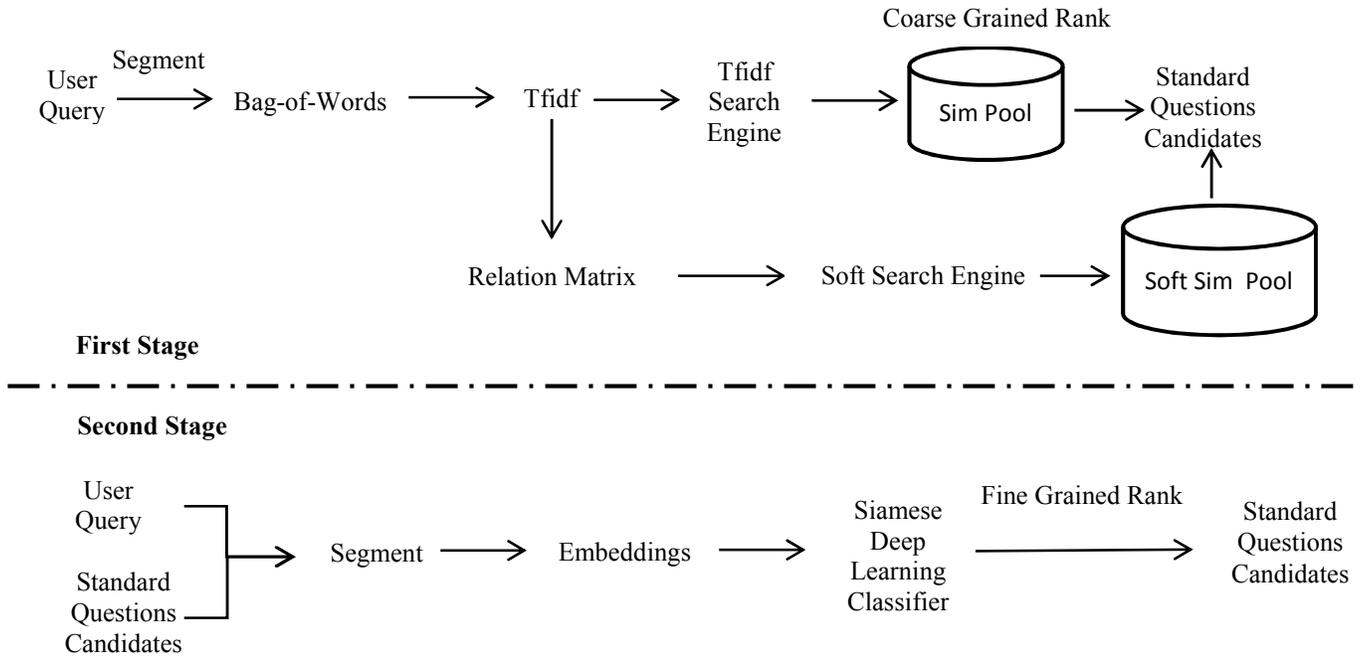

Fig 1. Two-stage framework

structure for the duplicate question pair classification problem.

### III. TWO STAGES NEURAL MATCHING MODEL

In this section, we present our model referred as two-stage neural matching model, shown in Fig. 1. The dash line seprates our model into a coarse-grained ranking stage and a fine matching phase.

#### A. Model Overview

Our model contains two stages as follows:
- We construct retrieval search engine integrated with similarity matrix for the coarse-grained ranking system;
- Subsequently, we employ a Siamese neural network with shared RNN weights layer to learn the deep semantic interaction of the paired information for fine tuning the ranking board from the previous step

In the following sections, we will explain the technical details in our two stages model.

#### B. Soft-similarities retrieve techniques

In this section, we show how we implement query search engine techniques to retrieve top-n question candidates. Firstly, questions with various lengths need to be padded into a vector space with same size, which is called vector space modeling (VSM). In a classical bag-of-words approach, TF-IDF coefficients are applied to fix the given sentence to an M-dimensional vector where M is the size of vocabulary occurring in the whole corpus. Given the standard questions as the dataset $\Phi = \{\psi_i \epsilon [0,1]\}_{i=1}^m$, and the user query $q\epsilon[0,1]$, the aim of the searching process is to find k items in $\Phi$ that is closest in VSM distribution to q:

$$KNN(q) = \arg\min dis(q, \phi) \quad (2)$$

where $dis(q, \phi)$ is the distance between query question and a given standard question in the FAQ list. KNN(q) represents the task of finding the K closest standard questions from dataset to the query. We utilise the cosine distance base to compute $dis(q, \phi)$ that involves the inverse of cosine dissimilarity when computing the *dis*.

TF-IDF coefficients compute the distance between user query and standard FAQ questions, such as cosine similarities.

$$\cos(A, B) = \frac{A^t . B}{\sqrt{A^t . A}\sqrt{B^t . B}}$$

$$where\ A^t . B = \sum_{i=1}^{M} \alpha_i \beta_i \quad (3)$$

The equation (3) above computes the summation of multiplication of co-occurring words' TF-IDF while zero is given when there is no common words between text A and text B. The previously mentioned cosine similarities have drawbacks that no value is captured if text A and B have words with linguistically similar meaning but are not the same. We therefore introduce a method called soft-similarity retrieval techniques that resolves the aforementioned drawbacks. A new matrix M is inserted between A and B comparing with original cosine similarity equation (3), details can be seen as follows:.

$$\cos_M(A, B) = \frac{A^t . M . B}{\sqrt{A^t . M . A}\sqrt{B^t . M . B}}$$

$$where\ A^t . M . B = \sum_{i=1}^{n}\sum_{j=1}^{n} \alpha_i m_{i,j} \beta_i \quad (4)$$

where M is a relation matrix that is designed to capture the relations among all words in confusion format. They enable us to obtain relevant semantic relations between words, based on a simple similarity measure between the vector representations of these words.

In this work, M is derived by distributed representations of words that have seen a tremendous success of word embedding techniques such as word2vec and glove approaches. We apply word2vec distributed representation in our framework empowered by the word2vec toolkit. The work2vec training model applies the CBOW configuration that is learned on multi-million scale text data extracted from online doctor-user dialogue. Each word2ec vector has the dimension 200, and only the words with a minimal frequency of 5 are taken into account. Consequently, a word2vec corpus containing 55411 candidate elements of term is constructed.

Once the word2vec representations of words are available, M can be computed in different ways. We have explored different variants, and the best results were obtained.

### C. Siamese neural networks

Long Short-Term Memory (LSTM) model was suggested by Hochreiter [28], as one of the recurrent neural networks(RNN), and it has made acknowledgeable successes in variable-length and time-series related inputs such as sentences process. Typical RNNs model can be simply interpreted as a standard neural network for sequence data $(x_1, \cdots, x_T)$ that updates to a hidden-state vector $h_t$ as follows:

$$h_t = sigmoid(Wx_t + Uh_{t-1}) \qquad (5)$$

Simple RNN is notorious on optimization difficulties due to the gradient vanish. The LSTM outperforms simple RNN model because the LSTM learns wide range dependencies through its use of memory cell units that selectively keeps information across variable-length input sequences. Concretely, the LSTM learns the hidden-state representation through four components: a memory state $C_t$, an output gate $O_t$, input gate $i_t$ and forget get $f_t$. Each t ∈ $\{1, \cdots, T\}$ in LSTM can be expressed as following equations with multiple weight matrices $W_i, W_f, W_c, W_o, U_i, U_f, U_c, U_o$ and bias-vectors bi, bf, bc, bo:

$$i_t = sigmoid(W_i x_t + U_i h_{t-1} + b_i) \qquad (6)$$

$$f_t = sigmoid(W_f x_t + U_f h_{t-1} + b_f) \qquad (7)$$

$$\tilde{c}_t = \tanh(W_c x_t + U_c h_{t-1} + b_c) \qquad (8)$$

$$c_t = i_t \odot \tilde{c}_t + f_t \odot c_{t-1} \qquad (9)$$

$$o_t = sigmoid(W_o x_t + U_o h_{t-1} + b_o) \qquad (10)$$

$$h_t = o_t \odot \tanh(c_t) \qquad (11)$$

Generally, LSTM [29] has been demonstrated to be well performed in various NLP tasks such as text modeling, sequence tagging and question answering system. Our proposed neural network is implemented as a concise Siamese LSTM neural structure [23] and is capable of interpreting domain knowledge customised features. It suits the problem that a set of pair sequences $(x_{s1}, x_{s2})$ and single label y are given. A paired sentences $(x_{s1}, x_{s2})$ is firstly padded to fixed-length vectors $(x_{s1}^a, x_{s2}^b \in \mathbb{R}^d)$. Customised features $x_f$ such as co-occurrence word counts are derived from operation between $x_{s1}$ and $x_{s2}$. The core network is built on the mutual-interaction among LSTM's last state of sequences $(x_{s1}, x_{s2})$ and domain features $x_f$. The network flow is as following: 1. Two representations are derived from the last state output of LSTM neural network. 2. The summation of the two representations is then generated. 3. Manhattan difference is applied in this work which uses the first order difference since Chopra [29] illustrates l1 norm outperforms the l2 norm. Unlike the Mueller [23] using single distance value, we use fixed size vectors to have better representation when considering the integration with other useful handcrafted feature vectors. 4. Mapping all features and sentences interactions by dense layers and anti-overfitting layers to the loss objectives. It is noteworthy that the left-handed side $LSTM_1$ shares all weights with right-handed side $LSTM_2$ due to the effectiveness of processing symmetric domains like sentence pairs. This special design empowers the network to capture the deep interaction of pair inputs in neural network dimension and domain knowledge dimension.

## IV. EXPERIMENT SETTINGS

### A. Data Set and Statistics

All training data are transformed from online hospital daily operational streams. Most of raw data is stored in the format of chat log (or text generated from automatic speech recognition) in Chinese that contains the full cycle of medical consultancy process including main healthcare concern question, diagnosis of question answering, and supplement question answering. Restricted by the policy of data protection and user privacy, data set is not allowed to go public. Alternatively, we slice an example and provide an overall statistic of our applied data set. By digging into the raw data-set distribution, we found a considerable share of question-answering pairs is general practice problem that can be responded from a pre-defined question-answering set. For the experiment purpose, we select one of the most popular department obstetrics & gynaecology as a representative example. We develop a question pool that contains 6414 standard questions and each question links to an answer compiled by many professional medical doctors. The pool contains 17 categories and is expected to cover frequent answered questions for the chosen department. The distribution is show in Fig. 2.

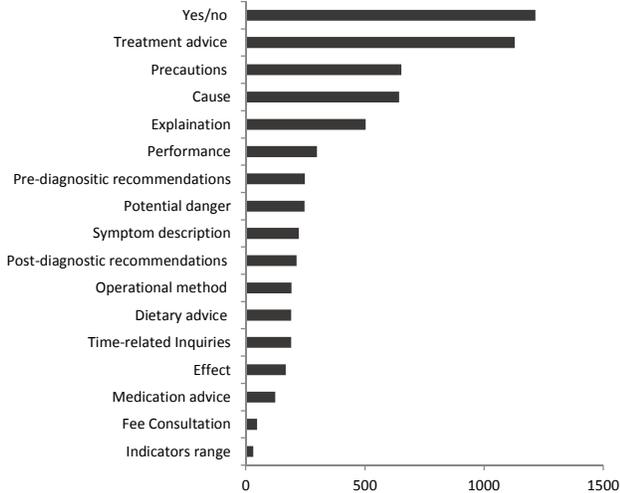

Fig. 2. Standard questions distribution

For establishing the information retrieval system, we use all questions in the pool to compute and store the searching index for IR. For training the deep neural network, over 160000 pairs of user questions and standard questions are labeled. The general data structure is presented as the following Table I such that user questions, standard questions and is able to respond are the three attributes.

TABLE I.  TRAINING QUESTION PAIR SAMPLES FOR DUPLICATE-QUESTION CLASSIFICATION

| Id | User questions | Standard questions | Is able to respond |
|---|---|---|---|
| 1 | What is the status of breast milk jaundice [母乳性黄疸是什么情况啊？] | How is breast milk jaundice diagnosed? [母乳性黄疸是如何诊断的呢？] | 1 |
| 2 | Is ok breast milk jaundice vaccination? [母乳性黄疸打预防针行不行？] | How is breast milk jaundice diagnosed? [母乳性黄疸是如何诊断的呢？] | 0 |
| 3 | What is the status of breast milk jaundice [母乳性黄疸是什么情况啊？] | What is neonatal jaundice? [新生儿黄疸是什么？] | 0 |

The training set, validation set distribution is 90%, 10% and the test set is an exclusive prepared 500 query-question pairs. We mark 1 for user question and standard question having similar meaning and therefore can link to same answer text, otherwise 0.

*B. Training Settings*

**Word Embedding**. We train our word embedding from scratch based on our large scale online data. Following the common practice in recent word2vec embedding models, we apply the configuration of window size 5 in Continuous Bag of Words (CBOW) model, and filtering the vocabulary less than occurring 5 times. In real training settings, only frequent word's vectors are stored and the rest words are assigned to a random sampled vector with uniform distribution sampling between [-0.1, -0.1].

**Hyper-parameters of our model**. For the setting of hyper-parameters, we set the word embedding dimensions as 200. We apply the N-Adams optimizer [30] in the settings of learning rate for 0.002 and beta1 for 0.9 and beta2 for 0.999 .

*C. Measurement Metrics*

MRR ratio is a commonly used evaluation metrics in information retrieval and question answering. In this work, we apply MRR to measure our rank answer sentences with the ground truth.

The definition of MRR is as following:

$$\text{Mean Reciprocal Rank} = \frac{1}{Q}\sum_{q=1}^{Q}\frac{1}{rank(first\ correct\ answer)} \quad (12)$$

where rank (first correct answer) is the place of the first occurring correct answer in the ranking list. The first correct answer is suitable for the question having only one correct answer.

V. EXPERIMENT RESULTS

*A. First Stage Information Retrieval*

The first baseline algorithm is ranking user query-standard questions directly by TFIDF coefficient search techniques. As can be seen from the black columns in Fig. 3, successful retrieve question counts and ratios are evaluated from retrieving 10 candidates to retrieving 100 candidates. When search candidates are limited to 10, only 245 out of 500 correct candidates are extracted where over half queries cannot be retrieved from this baseline method. As candidates number increases to 100, the successful extraction queries are rising to 397 that means nearly 80% queries are shown at least in top 100 candidates.

The second baseline is ranking our problems by soft similarity search techniques. The counts and ratios shown in blank columns present an overall outperforming performance than our base line method. For example as shown in Table II, one query says "Do I need surgery if I have my bone cleft? [骨裂的话要做手术吗？]" that would link to standard question "Do I need surgery after fracture? [骨折后需要手术吗?]". TFIDF search method fails to extract the right standard question in top 100 from question pools, however, soft search method ranks the correct candidate at the first place. This example implies soft search method has some extent of ability to link the semantic relevant words such as bone cleft [骨裂] and fracture[骨折]. Consequently, soft similarity method exerts 8%~20% higher performances than TFIDF search similarity methods, and less candidate numbers have generally greater improvements effects than more candidate numbers.

The best MRR of TFIDF search techniques and soft techniques are 0.2303 and 0.345 respectively. Intuitively, we can interpret the results the averages of ranking for our two search techniques are at position 4.3 and position 2.9 respectively.

TABLE II. EXAMPLES OF TFIDF SEARCH AND SOFT SEARCH

| Methods | Queries | Standard Questions | Rank |
|---|---|---|---|
| TFIDF Search | "Do I need surgery if I have my bone cleft? [骨裂的话要做手术吗？]" | "Do I need surgery after fracture? [骨折后需要手术吗？]" | Not available in top 100 results |
| Soft Search | "Do I need surgery if I have my bone cleft? [骨裂的话要做手术吗？]" | "Do I need surgery after fracture? [骨折后需要手术吗？]" | 1 |

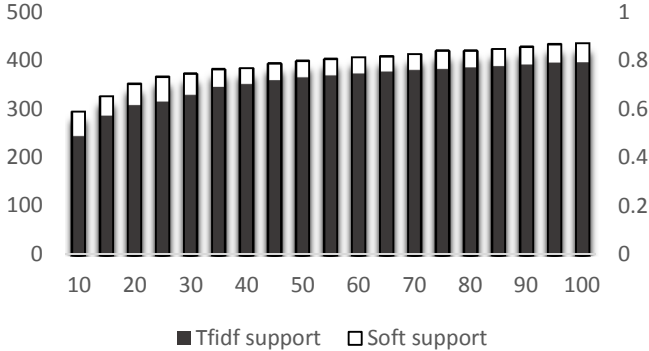

Fig. 3. TFIDF coefficient search engine counts (left y-axis) and ratios (right y-axis)

### B. Second Stage Information Retrieval

The third comparison algorithm is fine-grained re-ranking based on the first stage search engine results by second stage retrieving. Empowered by the deep learning for depth semantic understanding, the outputs of the first stage are the inputs of our deep learning model. The best MRR result of TFIDF +deep learning is 0.41 and the best MRR result of soft search techniques+deep learning is 0.44. Intuitively, our second stage framework can push the original mean position 4.3 and 2.9 forward to 2.4 and 2.3. In other words, the average chances of getting correct recommendation or answer text shows higher-ranking place when deep neural matching model is employed.

Fig. 4 shows the effects of changing the retrieving candidate numbers on our both two stage methods and both one stage baseline methods. For all two-stage scenarios, only at 10 candidates search scenario shows inferior performance than the one stage soft search method. This implies the first stage shows critical importance when successful retrieving candidates are low. As demonstrated in both Fig. 3 and Fig. 4, one stage TFIDF search techniques shows significant poor effects than soft search and has consistent low MRR value than its counterparts. This may drive two-stage margins between Soft-DL and TFIDF-DL methods along all scenarios but narrows down as retrieving candidates increase. Increasing candidates' number does not always enhance the MRR value for the second stage. Both tiidf+DL and soft+DL value hit the peak point when retrieving candidates set to 95, then goes down when candidate set to 100. Too many candidates may disturb the DL to rank the correct candidate to the top. In sum, those four curves illustrate the re-rank of second stage improves the first stage results and compensate the poor results of TFIDF search in comparison of Soft search as the first stage.

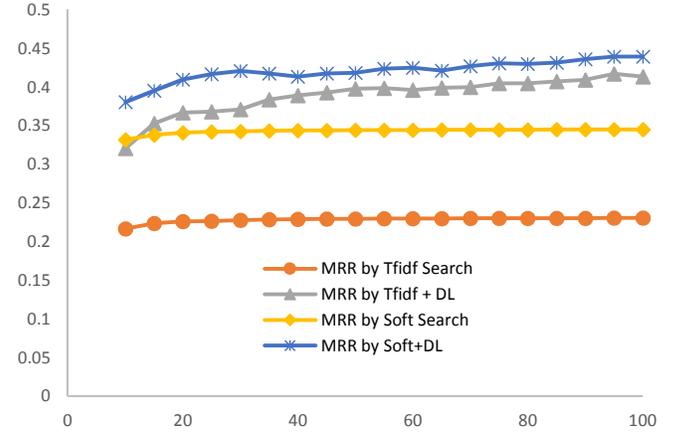

Fig.4 MRR ratio on different retrieving candidates (x-axis) via two one-stage methods (TFIDF Search and Soft Search) and two two-stage methods (TFIDF Search+Deep Learning and Soft Search+Deep Learning). Ratios on the y-axis are MRR value accordingly

### VI. CONCLUSIONS

Our work combines the traditional information retrieval technique and deep learning methods as a holistic two-stage model to efficiently retrieve and rank the short medical questions. Medical online queries, as our main processing contexts, are mainly in dialogue formats that have the nature of oral languages informality. The input data preserves the nature of ambiguity, arbitrary and fraction, which brings much challenge to NLP for retrieving the structure of user question against the pre-defined standard questions pool. We address those issues by a two-stage framework with the first stage coarse-grained shortlisting and subsequently the second stage fine-grained ranking model. The first stage of the proposed framework applies the plain TFIDF search and soft search to generate a primitive list of standard question candidates. The second stage uses the deep learning method to re-rank the candidates that can optimise the first stage results. The experimental results show that our two-stage framework achieves the best MRR rate 0.45 across the competitors. Future work includes the investigation of a more comprehensive method to rank the candidates focusing not only on the standard questions pool but also from the perspective of corresponding answer to the standard questions.

ACKNOWLEDGEMENTS

This work is partly supported by the National Natural Science Foundation of China (Grant No. 71471092), Natural Science Foundation of Zhejiang Province (Grant No. LR17G010001) and Ningbo Municipal Bureau of Science and Technology (Grant No. 2014A35006, 2017D10034)